\documentclass[letterpaper]{article} 
\usepackage{aaai2026}  
\usepackage{times}  
\usepackage{helvet}  
\usepackage{courier}  
\usepackage[hyphens]{url}  
\usepackage{graphicx} 
\usepackage{natbib}  
\usepackage{caption} 
\usepackage{algorithm}
\usepackage{algorithmic}
\usepackage{amsmath}
\usepackage{amsfonts}

\usepackage{xcolor}
\usepackage{booktabs}
\usepackage{multirow}
\usepackage{multicol}
\usepackage[textsize=tiny]{todonotes}  
\usepackage{newfloat}
\usepackage{listings}
\DeclareCaptionStyle{ruled}{labelfont=normalfont,labelsep=colon,strut=off} 
\floatstyle{ruled}
\newfloat{listing}{tb}{lst}{}
\floatname{listing}{Listing}
\nocopyright 

\usepackage{cleveref}

\newcommand{\methodname}{\texttt{AgREE}}
\newcommand{\changed}[1]{#1}

\title{AgREE: Agentic Reasoning for \\ Knowledge Graph Completion on Emerging Entities}

\author {
    Ruochen Zhao\textsuperscript{\rm 1},
    Simone Conia\textsuperscript{\rm 2},
    Eric Peng\textsuperscript{\rm 1},
    Min Li\textsuperscript{\rm 1},
    Saloni Potdar\textsuperscript{\rm 1}
}
\affiliations {
    \textsuperscript{\rm 1}Apple,
    \textsuperscript{\rm 2}Sapienza University of Rome\\
    \{ruochen\_zhao, ericpeng, s\_potdar, min\_li6\}@apple.com, simone.conia@uniroma1.it
}

\usepackage{bibentry}

\begin{document}

\maketitle

\begin{abstract}
Open-domain Knowledge Graph Completion (KGC) faces significant challenges in an ever-changing world, especially when considering the continual emergence of new entities in daily news.
Existing approaches for KGC mainly rely on pretrained language models' parametric knowledge, pre-constructed queries, or single-step retrieval, typically requiring substantial supervision and training data. Even so, they often fail to capture comprehensive and up-to-date information about unpopular and/or emerging entities. To this end, we introduce Agentic Reasoning for Emerging Entities (\methodname{}), a novel agent-based framework that combines iterative retrieval actions and multi-step reasoning to dynamically construct rich knowledge graph triplets. Experiments show that, despite requiring zero training efforts, \methodname{} significantly outperforms existing methods in constructing knowledge graph triplets, especially for emerging entities that were not seen during language models' training processes, outperforming previous methods by up to 13.7\%. Moreover, we propose a new evaluation methodology that addresses a fundamental weakness of existing setups and a new benchmark for KGC on emerging entities. Our work demonstrates the effectiveness of combining agent-based reasoning with strategic information retrieval for maintaining up-to-date knowledge graphs in dynamic information environments.\footnote{Our code and benchmark will be released upon acceptance.}
\end{abstract}


\section{Introduction}


\changed{Knowledge Graphs (KGs) are structured representations of entities and their semantic relationships, typically organized as collections of interconnected triples in the form of (subject, predicate, object)~\cite{hogan2021knowledge}. These graph-based data structures capture real-world knowledge by encoding entities as nodes and relationships as edges, enabling machines to understand and reason about complex interconnections in data~\cite{ji2021survey}. KGs have become essential infrastructure for numerous NLP applications, including question answering~\cite{kolomiyets2011survey}, information retrieval~\cite{xiong2017explicit}, and recommendation systems~\cite{wang2019kgat}.}

These knowledge repositories face the important task of incorporating the constant emergence of new entities in our dynamic world~\citep{emerging-entities}. We define an entity to be emerging if it has not been included in the knowledge graph before. For example, in real life, news sources regularly introduce previously undocumented entities such as emerging startups, rising public figures, novel products, newly-announced movies or TV-shows. This continuous influx of information creates a persistent gap between the static knowledge captured in existing KGs and the evolving real-world landscape. Therefore, how to dynamically update and expand the KG in a timely manner is an important problem to address.


Traditional KG completion methods relying on Knowledge Graph Embedding (KGE) models trained on the training corpus~\citep{Dettmers_Minervini_Stenetorp_Riedel_2018}, pre-trained language models~\citep{kgt5-context}, or even single-step retrieval~\citep{jiang-etal-2023-text} achieve good performances at predictions on previously established entities and relations. However, they could struggle with the incorporation of entirely new entities that lack any initial representation. This limitation significantly impacts the utility of KGs for applications that require up-to-date information.

Recent approaches~\citep{zhou-etal-2024-cogmg} have attempted to address this challenge by leveraging Large Language Models (LLMs) for new entity integration. However, these methods suffer from critical limitations. First, approaches that rely solely on LLMs' parametric knowledge are constrained by training cutoff dates, making them inherently outdated for truly unseen entities. Second, methods employing pre-constructed queries often fail to cover the diverse and unpredictable information spectrum surrounding new entities. Third, single-step retrieval approaches typically capture only a narrow slice of information, missing the multifaceted nature of entity relationships that comprehensive KG representation demands.

In this paper, we introduce \methodname{}, an agent-powered framework that combines strategic search and action planning to dynamically construct KG triplets for emerging entities. Unlike previous approaches, \methodname{} does not necessarily rely on static internal knowledge or predefined query templates. Instead, we employ an iterative agent-based exploration strategy that can strategically employ a set of tools to explore the web, self-reflect to plan the strategy and reason to complete KG triplets especially for emerging entities.

We perform extensive experiments to prove the effectiveness of our method. Evaluating on two commonly used Knowledge Graph Completion (KGC) datasets with hits@N, \methodname{} shows performance improvements up to 13.7\% on hits@N compared to previous models that are full-trained on the entire training set. To simulate a setup on emerging entities, we further construct a dataset of emerging entities by sampling the latest entities after a selected cutoff date. Due to the open-exploration ability, \methodname{} displays advantage on this dataset, showing up to 45.3\% hits@N improvements.
During performance evaluation, we also discover potential evaluation bias in traditional hits@N metric, and thus propose relation-based hits@N to account for 1-to-N relationships more fairly. When evaluating with relation-based hits@N, we observe even greater performance gaps compared to previous baselines, ranging from 21.8\% to 60.5\%. Thus, the method effectively integrates completely unseen entities that have never appeared in existing knowledge sources.

Our contributions can be summarized below:

\begin{enumerate}
    \item We introduce \methodname{}, a novel dynamic agent architecture for open domain knowledge completion and verification.
    \item Specifically, the framework utilizes multi-step retrieval and reasoning that progressively builds \changed{entity-centric subgraphs} through strategic information seeking.
    \item We perform comprehensive experiments and ablations, demonstrating consistent performance improvements in accuracy and ranking of the predictions on the KGC task compared to existing methods.
    \item We identify shortcomings with the current hits@N evaluation approach and introduce a relation-based hits@N metric with the aim of evaluating KGC task performance more fairly.
    \item We construct and release an emerging-entities dataset to better evaluate task performance on unseen entities.
\end{enumerate}

\section{Related work}



Previous research has established the importance of identifying and integrating emerging entities (EEs) into knowledge graphs. \citet{emerging-entities} formalized the definition of Emerging Entities or out-of-knowledge-base (OOKB) entities as those absent from existing knowledge bases. 
The WNUT2017 shared task \citep{derczynski-etal-2017-results} demonstrated the challenges in recognizing novel entities across diverse contexts. 
These works underscore the significance of our research motivation.

\changed{KGC methods has evolved through several methodological paradigms, from embedding-based approaches like TransE~\citep{bordes2013translating}, TuckER~\citep{balavzevic2019tucker}, and ConvE~\citep{Dettmers_Minervini_Stenetorp_Riedel_2018} that learn vector representations of entities and relations, to hybrid methods such as JointNRE~\citep{han2018neural} and RC-Net~\citep{xu2014rc} that incorporate textual information. Other works has explored reinforcement learning formulations, with MINERVA~\citep{das2017go} and CPL~\citep{fu2019collaborative} treating KGC as sequential decision-making through multi-hop reasoning paths.}
Recent advances in KGC have been driven by language models. 
At a high level, \citet{lv-etal-2022-pre} surveyed pre-trained language model approaches for KGC tasks. 
\citet{saxena-etal-2022-sequence} proposes KGT5, which reformulates KGC as a sequence-to-sequence problem and trains a model for link predictions. 
Incorporating Large Language Models (LLMs), \citet{zhou-etal-2024-cogmg} proposed CogMG, a collaborative framework between LLM and KGs that can identify knowledge triplets that are not present in the KG during question-answering. 
While these approaches show good performance on standard benchmarks, they rely heavily on parametric knowledge from training, potentially limiting effectiveness on post-training emerging entities.

To address the limitations of purely parametric approaches, researchers have also explored retrieval-augmented methods.
TagReal \citep{jiang-etal-2023-text} augments language models with prompt mining and context retrieval to ground KGC in external knowledge, while KGT5-context \citep{kgt5-context} extends KGT5 with contextual information, including neighborhood triplets and entity descriptions. These approaches attempt to utilize textual information, which could help in processing emerging entities. However, they typically employ single-step retrieval with pre-constructed query templates that may not adequately capture the diverse information around the new entities comprehensively. Additionally, they generally require supervised training to generate effective retrieval queries, limiting their flexibility when encountering entirely novel entity types.

Meanwhile, agent-based \changed{and LLM-based} approaches have shown remarkable effectiveness in tasks requiring multi-step reasoning and information gathering.
\changed{On LLM agents}, ReAct~\citep{react} was a pioneering work that  demonstrated how LLMs can interleave reasoning traces with actions to solve complex tasks, while ToolLLM \citep{toolformer} enabled models to learn to use external tools through instruction tuning. 
\changed{For Retrieval-Augmented-Generation on LLMs, rStar~\citep{ma2004rstar} and RARE~\citep{tran2024rare} show promising performance of utilizing monte-carlo tree search in solving multi-step problems.}
Despite their success in other domains, these agent-based reasoning frameworks have not been fully explored for open domain KGC especially for emerging entities.

\section{Methodology}

\begin{figure}[t!]
    \centering
    \includegraphics[width=0.45\textwidth]{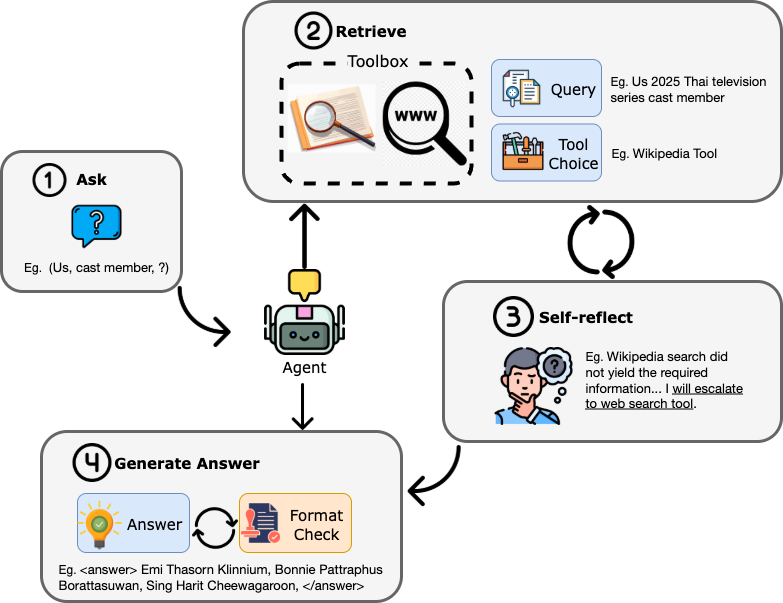}
  \caption{An illustration of the \methodname{} methodology.}
  \label{fig:method_fig}
\end{figure}

Our approach, \methodname, addresses the challenge of completing the information of emerging entities from external sources through an iterative agent-based framework that combines dynamic information retrieval with multi-step reasoning. Unlike previous approaches that rely on static parametric knowledge or single-step retrieval, our system employs a flexible, adaptive process that progressively builds comprehensive entity representations. An illustration of the overall method is provided in Figure \ref{fig:method_fig} with an illustrative example. A copy of the pseudo-code is provided in the supplementary materials.

As shown in Figure \ref{fig:method_fig}, as a first step, a query around a specific entity is given as input. In this example, the query is asking who the cast member is for ``Us''. \changed{Along with the query, we also feed the entity description (a Thai TV series)} into the agent, who is initialized with the task of predicting the answer to the query. \changed{If the entity is present in the existing KG, relevant triples are also shown to the agent to provide context.} Secondly, the agent determines whether to utilize tools (go to step 2) or to directly generate the answer (go to step 4) if it is already confident regarding its internal knowledge about the entity. This preliminary step prevents unnecessary retrieval operations for head entities while ensuring comprehensive information gathering for tail or unseen ones. In this example, the agent chooses to proceed to step 2 and looks up information with retriever tools.

In step 2, the agent utilizes the toolbox by producing a query and choosing a tool. In our framework, we provide two tools in the toolbox: basic retriever (powered by document retriever such as Wikipedia) and advanced retriever (powered by a web search API, such as Bing, Google). \changed{Different retriever tools provide coverage for different entities which can together provide better recall. Web search tools tend to cover more head entities whereas specific retrievers such as Wikipedia cover more medium and tail entities on a certain domain.} The agent is provided with descriptions of the tools \changed{and is encouraged to use the basic tool before escalation to the advanced tool.} In our example, the agent first calls the basic retriever with query ``Us 2025 Thai television series cast member''. 
Using these carefully constructed queries, \methodname{} retrieves information from external sources, obtaining a diverse set of documents containing potential knowledge about the emerging entity. The retriever here is a flexible plug-in component that can be substituted with other suitable retrievers for domain-specific tasks. 
\changed{To further improve information quality, we apply a three-stage process to the retrieved texts. First, we perform chunking to segment long documents into 3-sentence chunks. Second, we filter chunks with no overlapping keywords with the query to reduce irrelevant content. Finally, we re-rank the filtered chunks using the MS-MARCO-MiniLM model to prioritize the most relevant information. This procedure effectively mitigates noise from lengthy retrieved texts.}

After accessing the results of the tool, the agent proceeds to step 3, self-reflection, which is a key innovation in \methodname{}. The self-reflection mechanism evaluates the quality and sufficiency of the information retrieved. After each retrieval round, the agent analyzes the collected information and considers its completeness in predicting the current query result. This assessment determines whether additional information is needed. When the self-reflection indicates insufficient or imbalanced information, the agent initiates additional retrieval rounds with further improved queries or tool choices. These subsequent tool calls leverage insights from previous retrievals to target information gaps or resolve ambiguities. \changed{In this example, as the basic tool didn't yield information on cast members, the agent escalated to the advanced search tool with the same query.} This iterative refinement process continues until the retrieved information is deemed sufficient by the agent or reaches a maximum number of iterations, ensuring comprehensive coverage while maintaining computational efficiency. In this example, the agent reflects: ``Since the basic search did not yield the required information, I will escalate to the advanced search tool to retrieve more accurate and updated results.''. Then, it goes back to the toolbox and calls the advanced retriever with query.

In step 4, our system synthesizes the accumulated information into final answer predictions. The agent performs multi-step reasoning over the retrieved documents to identify supporting evidence and make the final prediction. After the answer is generated, we perform a rule-based format check to see whether the agent has followed the required answer format. If not, the generation step is repeated until a required format has been generated or a maximum number of iterations is reached. The rule-based check helps strengthen the format requirement. Finally, the produced answers are linked to entities with an entity-linking procedure. We limit the entire agentic process to a max number of iterations of 20. \changed{As a reference, we observe an average of 2.5 iterations for the Wikidata5m dataset.}

Overall, \methodname{} enables comprehensive knowledge graph expansion for emerging entities by combining the strengths of both LLMs and information retrieval systems in an agent-based framework. By interleaving retrieval with reflection and adaptation, \methodname{} predicts rich, accurate triplets around entities that have recently emerged and may not be well-represented in static knowledge sources.

\section{Experiments}
\label{sec:exps}

To evaluate \methodname{}, we conduct extensive experiments on commonly-used KGC datasets and compare against multiple baselines. Observing the experimental results, we discover a current shortcoming in the existing evaluation paradigms and propose to improve it with \textit{relation-aware hits evaluation}. To evaluate performances on newly emerged entities, we further propose a self-constructed Emerging-Entities dataset to highlight performance improvement on unfamiliar entities. Lastly, we perform ablation experiments to show the effectiveness of various components and present qualitative examples to analyze the method. 

\subsection{Experimental Setup}
\noindent\textbf{Implementation:} We implement \methodname{} using LangGraph, a toolkit designed for constructing LLM agent systems. 
By default, we use DeepSeek-V3~\citep{liu2024deepseek} as the backbone LLM, which is a 607B reasoning-specialized MOE model. 
For external information retrieval, we leverage two retrievers to access comprehensive and up-to-date knowledge: the basic retriever powered by the Wikipedia API and the advanced retriever powered by the Google search API. The agent has the freedom of choice to use the tools.\footnote{Prompts used are included in supplementary materials.}

\noindent\textbf{Datasets:} For evaluation, we employ two standard KGC datasets and construct an Emerging-Entities dataset. We include dataset statistics in the supplementary materials.

\begin{itemize}
    \item Wikidata5M~\citep{wang-etal-2021-kepler} comprises the 5 million most frequent entities from the Wikidata KG. Due to the prevalence of many-to-one relations in Wikidata5M, we focus solely on tail entity prediction.
    \item FB60K-NYT10~\citep{fu-etal-2019-collaborative} combines the Freebase KG with the New York Times text corpus. We evaluate both head and tail predictions for FB60K-NYT10.
    \item Emerging-Entities dataset covers entities created between after 2025.1, well after DeepSeek V3's training cutoff date to ensure the LLM has no parametric knowledge. The construction process is elaborated in the Experiments section.
\end{itemize}

\noindent\textbf{Baselines:} We compare \methodname{} with several approaches across different training paradigms for the KGC task.

\begin{itemize}
    \item \textbf{KGE-Based:} TransE~\citep{bordes2013translating}, TuckER~\citep{balavzevic2019tucker}, and ConvE~\citep{Dettmers_Minervini_Stenetorp_Riedel_2018} learn low-dimensional vector representations for entities and relations. TransE models relations as translations, TuckER employs tensor decomposition with a core tensor for multi-relational interactions, while ConvE uses 2D convolutions over reshaped embeddings for parameter sharing.
    
    \item \textbf{Text\&KGE-Based:} JointNRE~\citep{han2018neural}, RC-Net~\citep{xu2014rc}, and TransE+Line~\citep{fu2019collaborative} combine structured KG information with textual evidence. JointNRE uses joint neural frameworks for relation extraction and completion. RC-Net employs relation clustering with textual features. TransE+Line trains embeddings with LINE-based text representations through multi-task learning.

    \item \textbf{RL-Based:} CPL~\citep{fu2019collaborative} and MINERVA~\citep{das2017go} formulate KGC as sequential decision-making, using RL to navigate graphs. CPL focuses on collaborative policy learning across agents, while MINERVA uses single-agent curriculum learning with stochastic policies for path diversity.

    \item \textbf{PLM-Based:} TagReal~\citep{jiang-etal-2023-text}, PKGC~\citep{lv2022pre}, KGT5~\citep{saxena-etal-2022-sequence}, and SimKGC~\citep{wang2022simkgc} leverage pre-trained language models (PLMs) by converting KG triples into natural language formats. TagReal retrieves external textual evidence, PKGC uses KG-specific pre-training objectives, KGT5 treats KGC as sequence-to-sequence generation, while SimKGC employs contrastive learning with textual entity descriptions for KGC.

    \item \textbf{Context-Enhanced PLM:} KGT5-Context~\citep{kochsiek-etal-2023-friendly} builds upon KGT5's seq2seq approach by enriching input with entity descriptions and neighborhood subgraphs. This enhances understanding of entity semantics and local graph structure.

    \item \textbf{LLM-Agent:} ReAct~\citep{react} combines reasoning and acting in LLMs through an iterative framework.

\end{itemize}

\noindent\textbf{Metrics:} Following standard KGC evaluation practices, we employ Hits@N as our primary metric, which measures the proportion of test cases in which the ground truth entity appears within the top N predictions. For tail entity prediction, Hits@N is defined as follows:
\begin{equation}
\text{Hits@N} = \frac{1}{|T|} \sum_{(h,r,t) \in T} \mathbb{I}[\text{rank}(t|h,r) \leq N]
\label{eq:hits}
\end{equation}
where $T$ is the test set, $I[\cdot]$ is the indicator function, and $rank(t\mid h,r)$ denotes the rank of the ground truth tail entity $t$ given head entity $h$ and relation $r$. In the case of head entity prediction, the rank would be $rank(h\mid r, t)$.

We also evaluate accuracy with Mean Reciprocal Rank (MRR), calculated as the average of the reciprocal ranks assigned to the correct entities across all test instances. For tail entity prediction, MRR is defined as:
\begin{equation}
\text{MRR} = \frac{1}{|T|} \sum_{(h,r,t) \in T} \frac{1}{\text{rank}(t|h,r)}
\label{eq:mrr}
\end{equation}

To promote evaluation fairness, we additionally introduce relation-aware Hits@N, detailed in a later Section.

\subsection{Knowledge Graph Completion Performance}

\begin{table}[t!]
  \centering
\scalebox{0.65}{\begin{tabular}{ll|ccccc}
    \toprule
    Type & Method & Hits@1 & Hits@5 & Hits@10 & MRR \\
    \toprule
    \multirow{3}{*}{KGE} & TransE~$^*$ & - & 42.5\% & 46.8\% & 29.9 \\
    & ConvE~$^*$ & - & 50.2\% & 54.1\% & 40.4 \\
    & TuckER & 29.4\% & 48.8\% & 53.1\% & 38.6 \\
    \midrule
    \multirow{3}{*}{Text\&KGE} & JointNRE~$^*$ & - & 42.0\% & 47.3\% & 32.7 \\
    & RC-Net~$^*$ & - & 14.7\% & 16.3\% & 14.4 \\
    & TransE+Line~$^*$ & - & 26.8\% & 31.7\% & 11.0 \\
    \midrule
    \multirow{2}{*}{RL} & CPL~$^*$ & - & 43.3\% & 49.5\% & 33.5 \\
    & MINERVA~$^*$ & - & 43.8\% & 44.7\% & 34.6 \\
    \midrule
    \multirow{2}{*}{PLM} & TagReal & 12.4\% & 32.5\% & 48.5\% & 22.9 \\
    & PKGC~$^*$ & - & 42.0\% & 52.6\% & 32.1 \\
    \midrule
    \multirow{1}{*}{LLM-Agent} & \methodname{} (Ours) & \textbf{33.1\%} & \textbf{51.2\%} & \textbf{54.8\%} & \textbf{41.0} \\
    \bottomrule
    \multicolumn{6}{l}{$^*$Results taken from \citet{jiang-etal-2023-text}} \\
  \end{tabular}}
    \centering
  \caption{Performance comparison on FB60K-NYT10.}
  \label{table:fb60k-result}
\end{table}

\begin{table}[t]
  \centering
\scalebox{0.65}{\begin{tabular}{ll|cccc}
    \toprule
    Type & Method & Hits@1 & Hits@3 & Hits@10 & MRR \\
    \toprule
    \multirow{2}{*}{PLM} & SimKGC & 59.7\% & 67.6\% & 74.8\% & 65.0 \\
    & KGT5 & 49.0\% & 58.3\% & 67.3\% & 55.2 \\
    \midrule
    \multirow{2}{*}{Context\&PLM} & KGT5-Context & 37.3\% & 46.5\% & 54.7\% & 43.2 \\
    & KGT5-Context + Desc & 62.7\% & 71.2\% & 75.3\% & 67.1 \\
     \midrule
    \multirow{6}{*}{LLM-Agent} & \methodname{} (Ours) & \textbf{67.2\%} & \textbf{81.1\%} & \textbf{89.0\%} & \textbf{74.8} \\
    & \hspace{1ex} - Basic Retriever & 64.5\% & 78.7\% & 86.7\% & 72.3 \\
    & \hspace{1ex} - Advanced Retriever & 63.7\% & 77.8\% & 85.8\% & 71.4 \\
    & \hspace{1ex} - No self-reflection & 65.4\% & 87.1\% & 88.6\% & 73.0 \\
    & \hspace{1ex} - Mistral-Nemo & 30.5\% & 38.2\% & 40.0\% & 34.4 \\
    & \hspace{3ex} - no self-reflection & 34.6\% & 43.0\% & 45.3\% & 38.9 \\
    \bottomrule
  \end{tabular}}
    \centering
  \caption{Performance comparison on Wikidata5m dataset.}
  \label{table:wikidata-results}
\end{table}

\begin{table*}[t]
\centering
  \scalebox{0.7}{  \begin{tabular}{p{3cm}p{21cm}}
    \toprule
    Query & (Joseph W. Oxley, occupation, \textcolor{blue}{lawyer}) \\
    Predicted Entities & politician, sheriff, judge, \textcolor{blue}{lawyer}, ... \\
    Supporting Text & Joseph W. Oxley is an American Republican Party \textit{politician} who served as \textit{Sheriff} of Monmouth County, New Jersey from 1996 to 2007 ... \\
    \midrule
    Query & (Tomoaki Makino, member of sports team, \textcolor{blue}{Japan national football team}) \\
    Predicted Entities & Urawa Red Diamonds, \textcolor{blue}{Japan national football team}... \\
    Supporting Text & In Jan. 2012, he returned to J. League, loaned by 1. FC Köln to \textit{Urawa Red Diamonds}... He has represented the Japan national team internationally...  \\
    \midrule
    Query & (Maryana Marrash, sibling, \textcolor{blue}{Francis Marrash}) \\
    Predicted Entities & Abdallah Marrash, \textcolor{blue}{Francis Marrash} \\
    Supporting Text & Maryana Marrash Relatives: \textit{Francis Marrash} (brother); \textit{Abdallah Marrash} (brother)... \\
    \bottomrule
  \end{tabular}}
\caption{Examples of knowledge graph triplet completions that may not be fairly evaluated with hits@1.}
\label{table:rel-aware}
\end{table*}

We evaluate \methodname{} on two benchmarks and show the results in Tables \ref{table:fb60k-result} and \ref{table:wikidata-results}. In Table \ref{table:fb60k-result}, we include baselines from various training paradigms, including KGE-based, PLM-based, and RL-based. Specifically, the TagReal baseline is the most comparable to our method as it retrieves supporting evidence from a large text corpora to provide context for PLMs. Note that all the baselines in Table \ref{table:fb60k-result} are trained on the entire training set, whereas \methodname{} is completely training-free. As a result, we still observe higher performance in all metrics. 
\changed{We see that even though \methodname{}, as an LLM-Agent method, introduces more inference costs, it can be cost-efficient for applications with limited training data or when frequently updating knowledge bases, since it eliminates the computational overhead of retraining models on large-scale datasets.}

Similarly, by comparing against PLM-based and Context-enhanced PLM-based baselines on Wikidata5m, we observe consistent performance improvements on all hits metrics, ranging from 4.5\% to 13.7\%. The MRR metric is improved by 6.7. Thus, \methodname{} can achieve better KGC performance, despite having no training involved. Besides the main results, we include retriever analysis in supplementary materials to show how often the tools are called.

\subsection{Ablation Experiments}

In the main experiment, we equip the agent with two search tools: a basic retriever (powered by Wikipedia) and an advanced retriever (powered by Google search). As shown in the table, using only the basic retriever or only the advanced retriever results in degraded performance compared to using both tools together. Therefore, providing the combination of both retrieval tools yields the best performance, as it enables access to a wider range of information sources with complementary strengths. \changed{The improvement also shows that the agent can decide when to use each tool well.} When the self-reflection step is omitted, restricting the agent to always output the answer after an optional step of retrieval, the performance drops. Finally, we substitute the Deepseek-V3 backbone model with a smaller LLM, Mistra-Nemo 12B. The performance then drops significantly, indicating that the framework performance is dependent on LLM capabilities. As an interesting observation, we reduce the self-reflection step for Mistral-Nemo-based \methodname{} and observe that the performance actually improves. This shows that complex reasoning steps such as self-reflection could benefit models with strong capabilities, but the added complexity could harm less capable models. 

\subsection{Emerging Entities: A New Benchmark Dataset}
\label{subsec:ee_construction}

To evaluate model performance on truly unseen entities that emerged after training, we construct a benchmark dataset specifically targeting emerging entities\footnote{The code for dataset construction is included in our codebase.}. This dataset addresses a critical gap in existing KGC evaluation, as most benchmarks rely on entities that were likely encountered during model pre-training, introducing contamination.

\noindent\textbf{Entity Selection and Temporal Filtering:} We systematically select entities from Wikidata whose inception date falls between January 1, 2025 and May 1, 2025, which is well beyond DeepSeek-V3's training cutoff of July 2024. This temporal constraint ensures that the knowledge around the selected entities do not exist in the LLM's parametric knowledge, providing a rigorous test of generalization capabilities.

\noindent\textbf{Data Collection and Quality Control:} For each selected entity, we retrieve all associated triples via the Wikidata API. To maintain dataset quality and relevance, we apply two key filtering criteria: (1) \textit{Category filtering}: We restrict entities to 20 predefined categories including artists, music albums, TV shows, films, etc, focusing on domains with rich relational structure and public interest. (2) \textit{Cardinality filtering}: We exclude queries (h,r) with more than 10 ground-truth tail entities to avoid evaluation scenarios dominated by extensive entity lists that may dilute meaningful assessment.

\noindent\textbf{Dataset Statistics:} The resulting benchmark comprises 3,664 test triples spanning 90 distinct relations and 332 emerging entities. This scale provides sufficient diversity to evaluate model performance across various relation types while maintaining focus on the temporal emergence aspect. The dataset captures contemporary knowledge spanning entertainment, technology, and cultural domains that represent typical emerging entity scenarios in real-world applications.
Detailed dataset samples and comprehensive statistics are provided in supplementary materials.

\begin{table*}[htbp]
\centering
\scalebox{0.65}{
\begin{tabular}{|c|p{3cm}|p{22cm}|}
\hline
\textbf{Ex.} & \textbf{Triplet} & \textbf{Complete Trajectory} \\
\hline
\multirow{12}{*}{1} & \multirow{12}{3cm}{Us, cast member, \underline{Harit Cheewagaroon}} & 
\textbf{Step 0:} \underline{tool\_call}: basic\_search\_tool\\
& & \hspace{6ex} \underline{query}: "Us 2025 Thai television series cast member" \\
& & \hspace{6ex} \underline{Retrieved Results}: 2gether: The Series (Thai: ...; RTGS: Phr ... \\
& & \textbf{Step 1:} \underline{Self-Reflection}: The previous search provided information about various Thai television series and their cast members, but none of them directly matched the query for the 2025 Thai television series "Us."... I will escalate to the advanced search tool... \\
& & \textbf{Step 2:} \underline{tool\_call}: basic\_search\_tool\\
& & \hspace{6ex} \underline{query}: "Us 2025 Thai television series cast member" \\
& & \hspace{6ex} \underline{Retrieved Results}: Us: With Emi Thasorn Klinnium, Bonnie ... \\
& & \textbf{Step 4:} \underline{Self-Reflection}: The advanced search provided comprehensive and accurate information... \\
& & \textbf{Step 5:} \underline{Answer-Generation}: \textless answer\textgreater Emi Thasorn Klinnium, Bonnie Pattraphus Borattasuwan, Sing Harit Cheewagaroon, View Benyapa Jeenprasom, Pussarasorn Bosuwan, Thasorn Klinnium, Harit Cheewagaroon, Pattraphus Borattasuwan, Benyapa Jeenprasom, Thasorn Klinnium as Pam\textless /answer\textgreater \\
\hline
\multirow{8}{*}{2} & \multirow{8}{3cm}{Love Scout, genre, \underline{romantic drama}} & 
\textbf{Step 0:} \underline{tool\_call}: basic\_search\_tool\\
& & \hspace{6ex} \underline{query}: "Love Scout genre" \\
& & \hspace{6ex} \underline{Retrieved Results}: Love Scout (Korean: ...) is a 2025 South Korea ... \\
& & \textbf{Step 1:} \underline{Self-Reflection}: The previous search provided clear and accurate information about "Love Scout" as a 2025 South Korean romance drama workplace television series. The genre is explicitly mentioned as "romance drama"... there is no need for further retrieval. \\
& & \textbf{Step 2:} \underline{Answer-Generation}: \textless answer\textgreater romance drama, workplace drama, Korean drama, television drama, drama, romance, romantic drama, South Korean drama, workplace romance, romantic comedy\textless /answer\textgreater \\
\hline
\multirow{2}{*}{3} & \multirow{2}{3cm}{Battle of Baqubah, instance of, \underline{battle}} & 
\textbf{Step 0:} \underline{Answer-Generation}: ... Given that "Battle of Baqubah" is a military conflict, the most appropriate classification would be "battle" or "military conflict." ... \textless answer\textgreater battle, military conflict, armed conflict, event, historical event, war, Iraq War event, military operation, combat, engagement\textless /answer\textgreater \\
\hline
\end{tabular}%
}
\caption{\methodname{} Complete Trajectories: Three Qualitative Examples}
\label{tab:kg_agent_trajectories}
\end{table*}

\subsection{Performance on Emerging Entities}

\begin{table}[t]
  \centering
\scalebox{0.65}{\begin{tabular}{ll|cccc}
    \toprule
    Type & Method & Hits@1 & Hits@5 & Hits@10 & MRR \\
    \toprule
    Context\&PLM & KGT5-Context + Desc & 0.0\% & 0.0\% & 0.0\% & 0.0 \\
    PLM & SimKGC & 14.8\% & 18.7\% & 20.4\% & 16.9 \\
    \multirow{2}{*}{LLM-Agent} & ReAct & 30.2\% & 30.2\%  & 30.2\%  & 30.2  \\
    & \methodname{} (Ours) & \textbf{45.2\%} & \textbf{60.3\%} & \textbf{75.5\%} & \textbf{55.3} \\
    \bottomrule
  \end{tabular}}
    \centering
  \caption{Performance comparison on emerging entities dataset, including ablation variants.}
  \label{table:emerging-entities-result}
\end{table}

The results on our emerging entities dataset reveal significant performance disparities. KGT5-Context+Desc collapses completely on unseen entities, showing the brittleness of context-enhanced PLM approaches for entities absent from training data. SimKGC achieves modest performance (14.8\% Hits@1) through contrastive learning with textual descriptions. As KGC models fail on this task due to the limited access to new information, we also compare against ReAct~\citep{react}, which is an LLM Agent with the same backbone model and access to the wikipedia search tool. ReAct shows performance at 30.2\%, which is flat as it generates one answer. Our method substantially outperforms baselines with 45.2\% Hits@1, scaling to 75.5\% Hits@10, representing 15\% and 45\% improvements over ReAct respectively.

\subsection{Limitations of Current Evaluation: Introducing the Relation-Aware Hits@N Metric}
\label{subsec:ra_hits}

While the standard Hits@N evaluation metric provide useful benchmarks for KGC, they can produce misleading results when applied to relations with varying cardinalities. As illustrated in Table~\ref{table:rel-aware}, certain relations naturally admit multiple valid answers, making the choice of N in Hits@N evaluation potentially unfair for different relation types.
Consider the first example: Joseph W. Oxley holds multiple occupations (politician, sheriff, lawyer), yet standard Hits@1 evaluation would penalize a model for correctly predicting "politician" when the ground truth is ``lawyer.'' This evaluation bias is particularly problematic for one-to-many (1-N) relations, where entities can legitimately have multiple tail entities for the same relation type.

To address this limitation, we propose \textbf{relation-aware Hits@N}, which adapts the evaluation threshold based on each relation's inherent cardinality patterns observed in the training data. For each relation $r$, we define $N_{rel}$ as the maximum number of distinct tail entities associated with any head entity for relation $r$ in the training set. $\mathcal{E}$ is the set of all entities and $\mathcal{G}_{train}$ is the training KG:

\begin{equation}
N_{rel} = \max_{h \in \mathcal{E}} |{t : (h, r, t) \in \mathcal{G}_{train}}|
\label{eq:nrel}
\end{equation}

Our relation-aware Hits@N then evaluates each test triple (h,r,t) only when $N \leq N_{rel}$, ensuring that the evaluation criterion aligns with the relation's observed cardinality:

\begin{equation}
\text{Hits@N}_{RA} = \frac{1}{|T_{valid}|} \sum_{(h,r,t) \in T_{valid}} \mathbb{I}[\text{rank}(t|h,r) \leq N]
\label{eq:ra_hits}
\end{equation}

where $T_{valid} = \{(h,r,t) \in T : N \leq N_{rel}\}$ represents the subset of test triples for which the evaluation is meaningful given the relation's cardinality. While the scenario above shows tail entity prediction as an example, the same logic applies for head entity predictions by swapping $h$ and $t$ in Equation \ref{eq:nrel} and \ref{eq:ra_hits}.

This approach provides a more nuanced evaluation that respects the semantic properties of different relations, leading to fairer comparisons across models and more reliable assessment of KGC performance.

\subsection{Relation-Aware Hits Evaluation Results}

\begin{table}[t]
  \centering
\scalebox{0.65}{\begin{tabular}{ll|cccc}
    \toprule
    Dataset & Method & Hits@1 & Hits@3 & Hits@10 \\
    \toprule
    
    \multirow{2}{*}{FB60K-NYT10} & TagReal & - & 26.0\% & 56.2\%  \\
    & \methodname{} (Ours) & \textbf{87.1\%} & \textbf{92.0\%} & \textbf{92.9\%} \\
    \midrule
    \multirow{8}{*}{Wikidata5m} & KGT5-Context + Desc & 70.6\% & 60.4\% & 74.7\% \\
    & \methodname{} (Ours) & \textbf{82.4\%} & \textbf{93.6\%} & \textbf{93.4\%} \\
    & \hspace{1ex} - no self-reflection & 64.7\% & 87.1\% & 88.6\% \\
    & \hspace{1ex} - no entity-linking & 82.4\% & 93.1\% & 93.0\% \\
    & \hspace{1ex} - Basic Retriever & 70.6\% & 84.7\% & 86.6\% \\
    & \hspace{1ex} - Advanced Retriever & 70.6\% & 84.3\% & 84.2\% \\
    & \hspace{1ex} - Mistral-Nemo & 14.3\% & 37.0\% & 37.5\% \\
    & \hspace{3ex} - no self-reflection & 29.4\% & 37.5\% & 40.8\% \\
    \midrule
    \multirow{2}{*}{Emerging} & ReAct & 55.6\%  & 42.9\% & 25.0\% \\
    & \methodname{} (Ours) & \textbf{88.9\%} & \textbf{100.0\%} & \textbf{75.8\%} \\
    \bottomrule
  \end{tabular}}
    \centering
  \caption{Performance comparison for relation-aware evaluation on various datasets.}
  \label{rel-aware-result}
\end{table}

\begin{table*}[htbp]
\centering
\scalebox{0.65}{
\begin{tabular}{|c|p{4cm}|p{20cm}|}
\hline
\textbf{Error Type} & \textbf{Triplet} & \textbf{Trajectory Summary} \\
\hline
\multirow{2}{*}{\parbox{1.5cm}{Over-\\confidence}} & \multirow{2}{*}{\parbox{4cm}{Barbara Johnson, occupation, \underline{journalist}}} & 
\textbf{Step 0:} \underline{Answer-Generation}: Based on the known information, Barbara Johnson had multiple occupations throughout her life. Here are the identified occupations: ... $\langle$answer$\rangle$ literary critic, translator, philosopher, professor, writer, author, scholar, educator, academic, critic $\langle$/answer$\rangle$ \\
\hline
\multirow{7}{*}{\parbox{1.5cm}{Incomplete\\Retrieval}} & \multirow{7}{*}{\parbox{4cm}{Resident Playbook, cast member, \underline{Lee Bong-ryun}}} & 
\textbf{Step 0:} \underline{tool\_call}: basic\_search\_tool \\
& & \hspace{6ex} \underline{query}: ``Resident Playbook cast members'' \\
& & \hspace{6ex} \underline{Retrieved Results}: Resident Playbook (Korean: ...) is a 2025... \\
& & \textbf{Step 2:} \underline{Self-Reflection}: The search results provided comprehensive information about the cast members... including the main actors: Go Youn-jung, Shin Si-ah, Kang You-seok, Han Ye-ji, and Jung Joon-won... Since the information is sufficient, there is no need to escalate... \\
& & \textbf{Step 3:} \underline{Answer-Generation}: $\langle$answer$\rangle$ Go Youn-jung, Shin Si-ah, Kang You-seok, Han Ye-ji, Jung Joon-won, Bae Hyun-sung, Kim Dae-myung, Jung Kyung-ho, Park Bo-young, Park Jin-young $\langle$/answer$\rangle$ \\
\hline
\multirow{12}{*}{\parbox{1.5cm}{Irrelevant\\Retrieval}} & \multirow{12}{*}{\parbox{4cm}{Even Given the Worthless ``Appraiser'' Class, I'm Actually the Strongest, genre, \underline{adventure anime and manga}}} &  
\textbf{Step 0:} \underline{tool\_call}: basic search tool \\
& & \hspace{6ex} \underline{query}: ``Even Given the Worthless Appraiser Class, I'm Actually the Strongest genre'' \\
& & \hspace{6ex} \underline{Retrieved Results}: ...light novel series... manga adaptation... anime television series adaptation... [mixed with irrelevant voice actor and band information] \\
& & \textbf{Step 1:} \underline{Self-Reflection}: The search results provided information... However, the genre was not explicitly mentioned in the retrieved results. I will escalate to the advanced search tool to confirm the specific genre(s). \\
& & \textbf{Step 2:} \underline{tool\_call}: advanced search tool \\
& & \hspace{6ex} \underline{query}: ``Even Given the Worthless Appraiser Class, I'm Actually the Strongest genre'' \\
& & \hspace{6ex} \underline{Retrieved Results}: [...] Genre, Adventure, fantasy, harem [...] Genres: Action, Adventure, Fantasy \\
& & \textbf{Step 3:} \underline{Self-Reflection}: The advanced search results confirm that the genres [...] are Action, Adventure, and Fantasy... \\
& & \textbf{Step 4:} \underline{Answer-Generation}: $\langle$answer$\rangle$ action, adventure, fantasy, harem, light novel, Japanese anime, anime, shōnen, seinen, manga $\langle$/answer$\rangle$ \\
\hline
\end{tabular}%
}
\caption{Error Analysis: Common Failure Patterns and Trajectories}
\label{tab:kg_agent_errors}
\end{table*}

Table~\ref{rel-aware-result} presents the performance comparison relation-aware Hits@N evaluation across three benchmark datasets and on selected baselines. The support for evaluation sets are included in supplementary materials. The results demonstrate that our method consistently outperforms existing baselines. \changed{When evaluated more fairly with relation-aware hits@N, the performance margin is much more apparent compared to previous evaluation metrics in Table \ref{table:fb60k-result} and Table \ref{table:wikidata-results}, especially on ablations such as self-reflection}.

For FB60K-NYT10, our method demonstrates exceptional performance with 87.1\% Hits@1 and over 92\% for both Hits@3 and Hits@10, substantially outperforming TagReal across all metrics.
On Wikidata5m, \methodname{} achieves improvements over the strongest baseline (KGT5-Context + Desc), with gains ranging from 11.8\% to 33.2\% in Hits. 
Ablation studies show that self-reflection contributes significantly to performance, with its removal causing a 17.7\% drop in Hits@1.
On our Emerging entities dataset, we could observe the greatest performance improvement of \methodname{}. The results show performance gaps ranging from 32.7\% to 51.1\% compared to ReAct, demonstrating our method's effectiveness in handling never-seen entities.

Specifically, in Table \ref{table:rel-aware}, almost all hits metrics range from 80\% to 90\%, indicating that \methodname{}'s strength lies not only in identifying single correct answers but also in capturing the full spectrum of valid completions for relations with multiple legitimate targets, providing a more comprehensive assessment of KGC capabilities.

\subsection{Qualitative Examples and Error Analysis}

The trajectories in Table~\ref{tab:kg_agent_trajectories} demonstrate \methodname{}'s adaptive retrieval strategy across different knowledge completion scenarios. The first two examples are from the emerging entities dataset, whereas the last example is from wikidata5m dataset. Example 1 illustrates the agent's escalation mechanism: when the basic search for "Us" cast members returns irrelevant results about unrelated Thai series, the agent recognizes the insufficient information quality during self-reflection and automatically escalates to advanced search, ultimately retrieving comprehensive cast information. In contrast, Example 2 demonstrates efficient single-step retrieval, where the basic search immediately provides accurate genre information for "Love Scout," leading the agent to conclude the search without unnecessary additional calls. Most notably, Example 3 reveals the agent's capacity for knowledge-based reasoning without external retrieval—recognizing that the "instance of" relationship for "Battle of Baqubah" as a battle is well-established in its training knowledge, the agent bypasses retrieval entirely and directly provides taxonomic classifications. This adaptive behavior pattern, which escalates when needed, stops when sufficient, and avoids retrieval when unnecessary, demonstrates \methodname{}'s intelligent resource utilization and contextual reasoning capabilities, leading to both accurate knowledge completion and computational efficiency.

Moreover, we include an error analysis in Table~\ref{tab:kg_agent_errors} and reveal three distinct failure modes that highlight both the strengths and limitations of \methodname{}'s adaptive retrieval strategy. In the first example, the agent is over-confident and provides answer without retrieval despite having incomplete knowledge. The second example shows a different failure pattern where the agent performs retrieval but prematurely concludes the search after finding main cast members, missing supporting actors like "Lee Bong-ryun" that appear in the ground truth. Paradoxically, the agent expresses high confidence (``comprehensive information'', ``no need to escalate'') despite the incomplete coverage, suggesting limitations in the retrieval quality assessment mechanism. The last example highlights a failure pattern where the retrieved information is of low quality. In this case, when the initial basic search returns tangential information about voice actors and bands rather than genre classifications, the agent correctly identifies the inadequacy and escalates to advanced search. However, the advanced retriever still fails to return high-quality information on the genre that we are looking for. These findings indicate that improving \methodname{}'s calibration, particularly its ability to assess information completeness and trigger retrieval when internal knowledge is insufficient, represents a key area for future development.

\section{Conclusions}
In this paper, we introduced \methodname{}, an agent-based framework that addresses the challenge of open-domain KGC for emerging entities. By combining iterative retrieval with multi-step reasoning, our approach demonstrate substantial improvements over state-of-the-art baselines, representing a promising direction for maintaining up-to-date knowledge graphs in rapidly evolving information environments.

\bibliography{aaai2026}

\appendix

\section{Pseudo-Code}

\begin{algorithm}[htbp]
\begin{algorithmic}[1]
\REQUIRE Query $q=(e, r, ?)$ about entity $e$ and relation $r$; Tool reserve $\mathcal{T} = \{T_{basic}, T_{advanced}\}$
\ENSURE Predicted entity for query $(e, r, ?)$
\STATE Initialize agent with task of answering query $q$
\STATE $retrieved \leftarrow \emptyset$
\STATE $max\_iterations \leftarrow M$
\STATE $iteration \leftarrow 0$
\WHILE{$iteration < max\_iterations$}
    \STATE $iteration \leftarrow iteration + 1$
    \IF{agent confident about internal knowledge of $e$}
        \STATE go to Step 4 
    \ENDIF
    \STATE $search\_query \leftarrow$ agent.get\_query($retrieved$, $q$)
    \STATE $tool \leftarrow$ agent.select\_tool($\mathcal{T}$, $q$, $retrieved$)
    \STATE $new\_docs \leftarrow$ tool.retrieve($search\_query$)
    \STATE $retrieved \leftarrow retrieved \cup new\_docs$
    \STATE $sufficient \leftarrow$ agent.evaluate\_sufficiency($retrieved$, $q$)
    \IF{$sufficient$} 
        \STATE break 
    \ENDIF
    \STATE agent.reflect\_on\_gaps($retrieved$, $q$)
\ENDWHILE
\STATE $max\_gen\_attempts \leftarrow N$; $gen\_attempt \leftarrow 0$
\WHILE{$gen\_attempt < max\_gen\_attempts$}
    \STATE $gen\_attempt \leftarrow gen\_attempt + 1$
    \STATE $answer \leftarrow$ agent.synthesize\_answer($retrieved$, $q$)
    \IF{rule\_based\_format\_check($answer$)} 
        \STATE break 
    \ENDIF
\ENDWHILE
\STATE $linked\_entities \leftarrow$ entity\_linking($answer$)
\RETURN $linked\_entities$
\end{algorithmic}
\caption{\methodname{}}
\label{alg:methodname}
\end{algorithm}

\section{Prompts}
\label{app:prompts}

\subsection{Task Instruct Prompt}

You are a Knowledge Graph Completion Agent. Complete the missing entity in knowledge graph triplets (head\_entity, relation, tail\_entity).

TASK:

- When given head entity → find tail entity

- When given tail entity → find head entity

RELATION STRUCTURE:

- Relations consist of 3 words (format: domain/type/property)

- Example: people/person/birth\_date

- Second word (type) defines what the head entity should be (e.g., "person")

- Third word (property) defines what the tail entity should be (e.g., "date")

SPECIAL ENTITY HANDLING:

- If the target entity type is location, find the location on all possible granularities (e.g. town, city, state, country)

- If the relation can have multiple answers, such as "occupation" or "spouse" where a person could have multiple occupations/spouses throught the lifetime, try to find all potential answers and list out their associated time frames.

Information retrieval guidelines:

1. If you're confident in your knowledge, answer directly

2. Otherwise use retrieval tools in this order:

   - Start with search\_tool\_basic
   
   - Escalate to search\_tool\_advanced if needed
   
3. Always search thoroughly - even if you find one answer, make sure you haven't missed others

Do not ask for user input. Make decisions independently based on your knowledge and retrieved information.

[query]

Example triplets with relation [relation] (DO NOT REPEAT THESE TRIPLETS IN ANSWERS):

[triples with the same relation]

Known information about entity [entity] (DO NOT REPEAT THESE TRIPLETS IN ANSWERS):[entity description]

[triples with the same entity]

\subsection{Answer Generation Prompt}

ANSWER REQUIREMENTS:

- Provide at least 10 possible answers, ranked from most to least confident

- Include different granularity levels (e.g., if "tanker" is an answer, also include "ship")

- For location entities: include all granularity levels (neighborhood, town, city, state, country)

- Answer category and format should be similar to these examples: [endings of triples with the same relation]

OUTPUT FORMAT:

- STRICTLY use this format only: <answer>ANSWER\_1, ANSWER\_2, ...</answer>

- DO NOT include explanations, reasoning, or any text outside the <answer> tags

- Think concisely and provide answers immediately

\section{Dataset Statistics}
\label{app:dataset}

\begin{table}[H]
  \centering
\scalebox{0.65}{\begin{tabular}{l|ccccc}
    \toprule
    Dataset & Training Set & Validation Set & Test Set & Relations & Entities \\
    \toprule
    Wikidata5m & 4,579,609 & 7,374 & 7,475 & 828 & 4.8M \\
    FB60K-NYT10 & 268,280 & 8,795 & 8,918 & 1,253 & 37,530 \\
    Emerging-Entities & - & - & 3,664 & 90 & 332 \\
    \bottomrule
  \end{tabular}}
    \centering
  \caption{Benchmark Dataset Statistics.}
  \label{table:dataset-statistics}
\end{table}

We include basic dataset statistics in Table \ref{table:dataset-statistics}. It is worth noting that although the FB60K-NYT10 dataset covers 1,253 relations, it only retains 16 commonly-seen relations in the test set. Moreover, for FB60K-NYT10, we perform both head and tail entity predictions, whereas we only experiment with tail entity predictions for the other two datasets.

\subsection{Emerging Entity Dataset Construction}
\label{app:dataset_emerging}

\begin{table*}[t]
\centering
\scriptsize
\begin{tabular}{p{1.2cm}p{2cm}p{1cm}p{1.5cm}p{1.8cm}p{4cm}}
\toprule
\textbf{Entity ID} & \textbf{Entity Name} & \textbf{Inception Date} & \textbf{Description} & \textbf{Category} & \textbf{Triple} \\
\midrule
\multirow{3}{*}{Q133806770} & \multirow{3}{*}{Million Dollar Secret} & \multirow{3}{*}{2025-03-26} & \multirow{3}{*}{\parbox{1.5cm}{American television reality game show}} & \multirow{3}{*}{television series} & (Million Dollar Secret, filming location, Kelowna) \\
& & & & & (Million Dollar Secret, genre, Q111812458) \\
& & & & & (Million Dollar Secret, country of origin, United States) \\
\midrule
\multirow{3}{*}{Q132736363} & \multirow{3}{*}{Claude 3.7 Sonnet} & \multirow{3}{*}{2025-02-24} & \multirow{3}{*}{\parbox{1.5cm}{multimodal large language model}} & \multirow{3}{*}{\parbox{1.6cm}{large language model, reasoning language model}} & (Claude 3.7 Sonnet, developer, Anthropic) \\
& & & & & (Claude 3.7 Sonnet, instance of, reasoning language model) \\
& & & & & (Claude 3.7 Sonnet, part of, Claude) \\
\midrule
\multirow{3}{*}{Q127146791} & \multirow{3}{*}{Moonlight Mystique} & \multirow{3}{*}{2025-01-07} & \multirow{3}{*}{\parbox{1.5cm}{2025 Chinese net drama}} & \multirow{3}{*}{television series} & (Moonlight Mystique, original broadcaster, iQIYI) \\
& & & & & (Moonlight Mystique, cast member, Kent Tong) \\
& & & & & (Moonlight Mystique, genre, period drama film) \\
\midrule
\multirow{3}{*}{Q135078252} & \multirow{3}{*}{\parbox{2cm}{Hampton by Hilton Potsdam Babelsberg}} & \multirow{3}{*}{2025-04-15} & \multirow{3}{*}{Hotel in Potsdam} & \multirow{3}{*}{building, hotel} & (Hampton by Hilton Potsdam Babelsberg, industry, hospitality industry) \\
& & & & & (Hampton by Hilton Potsdam Babelsberg, country, Germany) \\
& & & & & (Hampton by Hilton Potsdam Babelsberg, owned by, Hilton Worldwide) \\
\midrule
\multirow{3}{*}{Q133812572} & \multirow{3}{*}{\parbox{2cm}{Art × Science International}} & \multirow{3}{*}{2025-04-05} & \multirow{3}{*}{\parbox{1.5cm}{non-profit organization based in France}} & \multirow{3}{*}{\parbox{1.6cm}{company, nonprofit organization, non-governmental organization, ...}} & (Art × Science International, founded by, Noé Cabannes Michel) \\
& & & & & (Art × Science International, official language, English) \\
& & & & & (Art × Science International, headquarters location, Rouen) \\
\bottomrule
\end{tabular}
\caption{Sample entities from the emerging entities dataset with their metadata and associated triples.}
\label{tab:emerging_entities_sample}
\end{table*}

We provide dataset statistics for the Emerging-Entities dataset in Table \ref{table:dataset-statistics} and Table \ref{tab:emerging_entities_sample}. Specifically, we selected 3 triples to show for each entity. The Emerging-Entities dataset covers a range of entities, spanning from TV shows to buildings. Along with the codebase, we will release code for constructing the dataset for evaluating in any period of time.

\subsection{Relation-Aware Evaluations}
\label{app:rel_aware}

\begin{table}[H]
  \centering
\scalebox{0.65}{\begin{tabular}{ll|ccc}
    \toprule
    Dataset & Testset Size & Hits@1 & Hits@3 & Hits@10 \\
    \toprule
    Wikidata5m & 5133 & 17 & 28 & 1375 \\
    FB60K-NYT10 & 17836 & 1909 & 2220 & 8738 \\
    Emerging-Entities & 3664 & 9 & 12 & 483 \\
    \bottomrule
  \end{tabular}}
    \centering
  \caption{Dataset Support for Relation-Aware Evaluations. Numbers shown are the number of data points evaluated for relation-aware hits@N in the test set.}
  \label{table:rel-aware-support}
\end{table}

Table~\ref{table:rel-aware-support} presents the coverage statistics for relation-aware evaluations reported in Table for relation-aware hits@N in the main paper. For the Emerging-Entities dataset, we leverage the Wikidata5m training set to determine the relation-specific cardinality $N_{rel}$ for each relation type. Our analysis reveals that both Wikidata5m and Emerging-Entities exhibit limited coverage, with relation-aware Hits@$N_{rel}$ applicable to only a restricted subset of test instances. This reduced coverage stems from the prevalence of one-to-many relations in Wikidata, where the high cardinality of certain relations results in $N_{rel}$ values that exceed practical evaluation thresholds. In contrast, FB60K-NYT10 demonstrates substantially broader coverage for relation-aware evaluation, as this dataset contains fewer one-to-many relations and more balanced relation cardinalities. Consequently, the relation-aware Hits@$N_{rel}$ metric can be meaningfully applied to a larger proportion of test cases in FB60K-NYT10, providing more comprehensive evaluation coverage.

\section{Retriever Calls in Experiments}
\label{app:retriever}

\begin{table}[H]
  \centering
\scalebox{0.65}{\begin{tabular}{l|c|c|c|c}
\toprule
\textbf{Dataset} & \textbf{Test Size} & \textbf{Basic} & \textbf{Advanced} & \textbf{Didn't Call} \\
& & \textbf{Retriever} & \textbf{Retriever} & \textbf{Retriever} \\
\midrule
Wikidata5m & 7,475 & 695 & 530 & 4,377 \\
FB60K+NYT10 & 17,836 & 8,548 & 5,846 & 8,091 \\
Emerging-Entities & 3,664 & 2,299 & 1,812 & 455 \\
\bottomrule
\end{tabular}}
\caption{Retriever usage statistics across different datasets.}
\label{tab:retriever_usage}
\end{table}

The retriever usage statistics in Table \ref{tab:retriever_usage} reveal distinct patterns between the datasets that reflect their inherent characteristics. On Wikidata5m, our method frequently relies on parametric knowledge without external retrieval (4,377 out of 7,475 cases, ~58.6\%), suggesting that the model's pre-trained knowledge sufficiently covers many entities in this established benchmark. When retrieval is needed, basic retrieval (695 cases) is preferred over advanced retrieval (530 cases), indicating that simpler retrieval strategies often suffice for well-documented entities. In contrast, the Emerging-Entities dataset shows dramatically different behavior, with heavy reliance on both basic (2,299 cases, ~62.7\%) and advanced retrieval (1,812 cases, ~49.5\%), while only 455 cases (~12.4\%) avoid retrieval entirely. This pattern demonstrates that emerging entities require substantial external knowledge acquisition, validating our dataset's effectiveness in testing models' ability to handle truly novel information beyond their training cutoff.

\end{document}